\documentclass[11pt]{article}

\usepackage[preprint]{acl}

\usepackage{times}
\usepackage{latexsym}
\usepackage{makecell}
\usepackage[T1]{fontenc}
\usepackage{tabularx}

\usepackage{array}

\usepackage[utf8]{inputenc}

\usepackage{microtype}

\usepackage{inconsolata}

\usepackage{graphicx}

\usepackage{booktabs} 
\usepackage{multirow}
\usepackage{tikz}

\makeatletter
\DeclareRobustCommand\onedot{\futurelet\@let@token\@onedot}
\def\@onedot{\ifx\@let@token.\else.\null\fi\xspace}

\def\eg{\emph{e.g}\onedot, }

\makeatother

\usepackage{listings}
\usepackage{svg} 
\usepackage{xspace}
\newcommand{\projname}{SWITCH\xspace}

\title{\projname: Benchmarking Modeling and Handling of Tangible Interfaces in   Long-horizon Embodied Scenarios}

\author{ 
    \textbf{Juntao Cheng}$^{1,2}$\thanks{Equal contributors.}\thanks{Work done during the internship at Beijing Academy of Artificial Intelligence (BAAI).},
    \textbf{Wanyue Zhang}$^{3,4}$\footnotemark[1],
    \textbf{Zhiwei Yu}$^{1}$ \thanks{Correspondence authors: <zwyu, borje>@baai.ac.cn}\thanks{Project lead.},
    \textbf{Shuo Ren}$^{4}$, \\
    \textbf{Zheqi He}$^{1}$, 
    \textbf{Shaoxuan Xie}$^{1}$, 
    \textbf{Guocai Yao}$^{1}$, 
    \textbf{Jieru Lin}$^{6}$\footnotemark[1],
    \textbf{Börje F. Karlsson}$^{1}$\footnotemark[3],
    \textbf{Jiajun Zhang}$^{3,4,5}$ \\
    $^{1}$Beijing Academy of Artificial Intelligence (BAAI)
    $^{2}$Shanghai Jiao Tong University\\
    $^{3}$School of Artificial Intelligence, University of Chinese Academy of Sciences \\
    ~$^{4}$Institute of Automation, Chinese Academy of Sciences \\
    ~$^{5}$Wuhan AI Research
    $^{6}$Harbin Institute of Technology \\
}

\begin{document}

\maketitle

\begin{tikzpicture}[remember picture, overlay]
    \node[anchor=north west, yshift=-0.4in, xshift=0.7in] at (current page.north west) {
        \includegraphics[width=3cm]{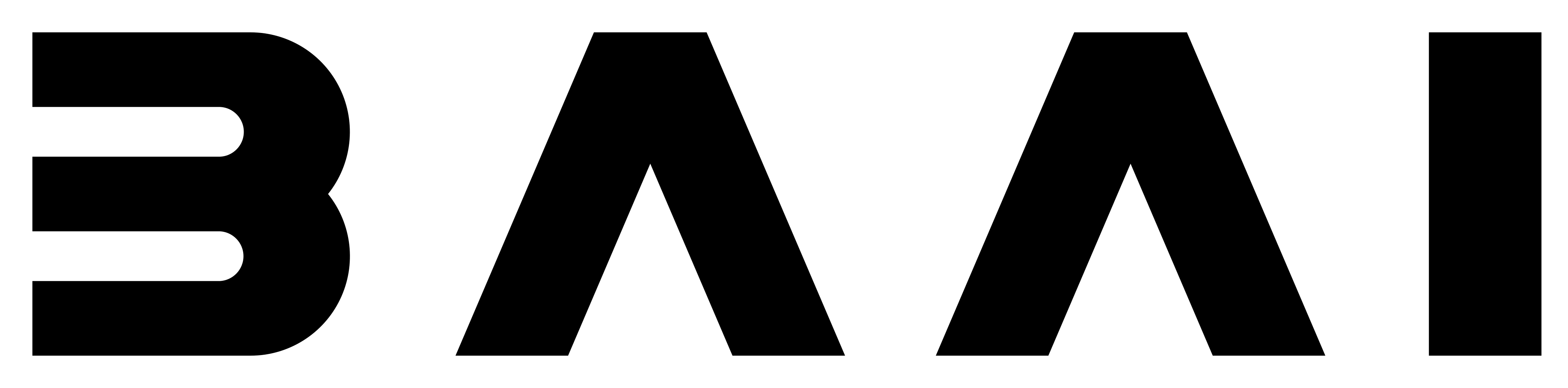}
    };
    
    \draw[darkgray] (current page.north west) ++(0.7in, -0.85in) -- ++(\columnwidth * 2.2, 0);
\end{tikzpicture}

\begin{abstract}
Tangible control interfaces (TCIs), such as appliance panels, remotes, elevators, and embedded GUIs, are a fundamental component of everyday human-built environments. Interacting with these interfaces requires agents not only to ground language in visual observations, but also to execute actions, track temporally evolving state changes, and verify whether intended outcomes have been achieved. 
However, existing benchmarks predominantly evaluate open-loop perception or single-step action execution, failing to capture this continuous cycle of interaction, feedback, and correction.  We introduce \textbf{\projname}, a benchmark for closed-loop interactive reasoning with TCIs in realistic egocentric environments\footnote{Benchmark resources: \url{https://huggingface.co/datasets/BAAI-Agents/SWITCH}}. \textbf{\projname} comprises 1,170 temporally interactive videos across diverse functional categories, providing structured annotations of instructions, actions, state transitions, outcomes, and recovery behaviors over time. 
To probe generative world modeling, \projname also evaluates video generation models on interaction-centered tasks using both LLM-as-judge and human evaluation\footnote{Leaderboard website: \url{https://huggingface.co/spaces/BAAI-Agents/SWITCH-Leaderboard}}. Experiments with frontier proprietary and open-source multimodal models reveal persistent weaknesses in fine-grained visual-temporal perception, outcome verification, and error recovery, highlighting \projname as a testbed for closed-loop embodied intelligence.
\end{abstract}

\section{Introduction}
\label{sec:intro}

\begin{figure*}[htbp]
\centering
\includegraphics[width=1\linewidth]{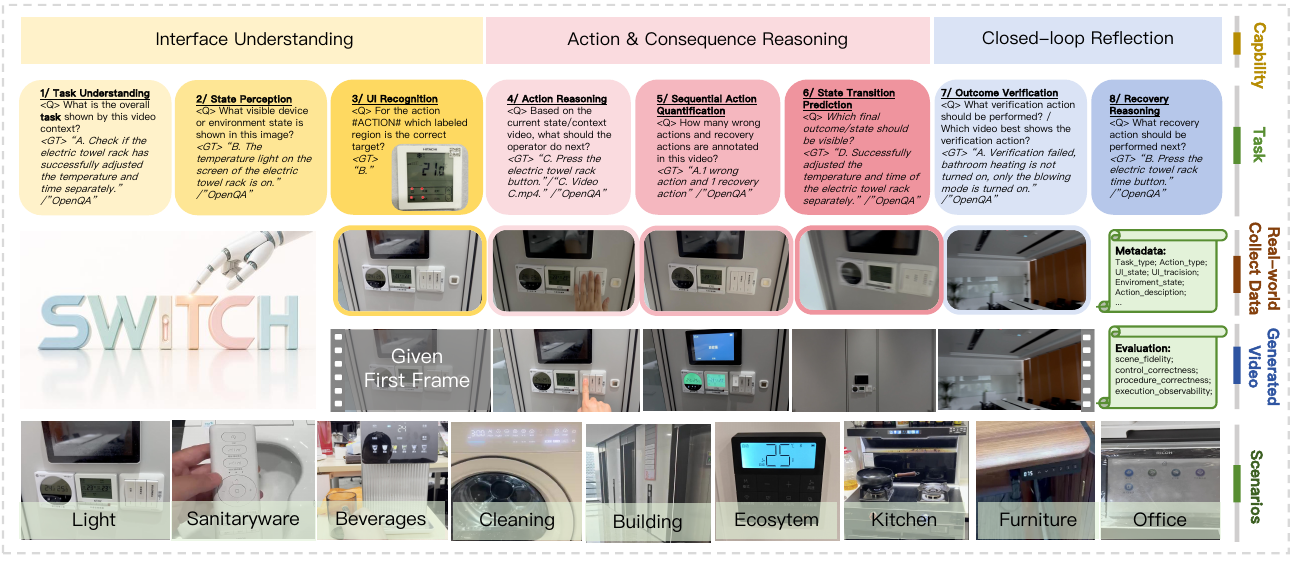}
\caption{
\textbf{Overview of the \projname benchmark.} 
The framework structures TCI interaction into \textbf{8 tasks} across \textbf{three core capabilities}: \textit{Interface Understanding}, \textit{Action \& Consequence Reasoning}, and \textit{Closed-loop Reflection}. 
We illustrate the pipeline using the example task ``Turn off all the lights'' and detail the progression from \textbf{Real-world Collected Data} to \textbf{Generated Video} evaluation. \projname covers 9 diverse domains spanning daily life and work environments (see Table~\ref{tab:dataset_scenarios} in the Appendices for details).
}
\label{fig:firstPic}
\end{figure*}

Intelligent behaviour in physical world requires not only perception and reasoning, but also effective interaction with the existing real world and the infrastructure embedded within it. In everyday human environments, this interaction is mediated largely through \textbf{tangible control interfaces (TCIs)}, including switches, appliance panels, elevators, and embedded GUIs. These interfaces form the primary bridge between human intentions and physical state changes in the environment.  
Agents must ground instructions in situated visual observations, infer interface affordances and device states, predict how actions will alter the environment, track temporally evolving state changes, and verify whether intended outcomes have actually been achieved. Unlike static perception tasks or open-loop action prediction, interaction with TCIs is inherently \textbf{closed-loop}: actions modify the world, observations evolve over time, and failures may require recovery or re-planning. Moreover, TCI interactions often involve delayed or spatially displaced effects—\eg preheating an oven, changing environmental controls, or activating a device outside the current field of view—requiring models to reason beyond immediate visual feedback. 

Despite substantial progress in computer vision, large multimodal models (LMMMs), and embodied agents, existing benchmarks  (\eg~\cite{foss2025causalvqaphysicallygroundedcausal, chen2025worldprediction, chow2025physbench}) only partially capture these challenges. An agent may select a plausible action but suffer from \textit{visual-semantic misalignment}: failing to recognize that the action did not achieve the goal, missing subtle visual state changes, or \textit{hallucinating task success} without recovering from an incomplete execution. This gap limits our ability to measure whether multimodal models can perform interactive reasoning rather than isolated perception or prediction.

We introduce \textbf{\projname} (\textbf{S}emantic \textbf{W}orld \textbf{I}nterface \textbf{T}asks for \textbf{C}ontrol \& \textbf{H}andling), a benchmark for evaluating interaction-centric capabilities and causality in the physical world. SWITCH formulates interface use as an interactive reasoning cycle consist of: (i) understanding instructions and interfaces, (ii) predicting next actions, (iii) tracking state transitions, (iv) verifying outcomes from visual feedback, and (v) recovering from failure through re-planning. \textit{This framing turns physical interaction into a multimodal reasoning problem centered on action consequences and feedback.}

SWITCH is built from 1,170 real-world egocentric interaction videos spanning nine major categories of everyday TCIs. It provides language-grounded temporal annotations of fine-grained actions, state transitions, task outcomes, and recovery behaviors. Based on these annotations, SWITCH supports three complementary formats: (1) Multiple-Choice QA (MCQ) for scalable automatic evaluation, (2) OpenQA for free-form reasoning and outcome verification, and (3) Interactive Video Generation for evaluating generative world modeling and interaction dynamics prediction. Figure \ref{fig:firstPic} illustrates the overall benchmark design.

Our contributions are threefold. First, we introduce \textbf{a real-world benchmark for TCI-centered closed-loop  interaction}, targeting a practical but underexplored setting for multimodal agents. Second, we provide \textbf{a unified evaluation protocol} spanning multiple-choice QA, open-ended rubric-based judging, temporal action reasoning, recovery reasoning, and generative video evaluation. Third, we \textbf{systematically evaluate frontier multimodal models}, revealing persistent weaknesses in fine-grained visual-temporal perception, outcome verification, and failure recovery, thereby charting a clear path for future research in aligning LMMMs with dynamic physical environments. Source code, datasets, and model inference results will be released publicly.



\section{Related Work}
Evaluating multimodal agents in realistic interactive environments remains challenging. Existing work mainly falls into two lines: physical/world-model evaluation and interface-control evaluation. Benchmarks such as PhysBench, DeepPHY, WorldPrediction, WorldBench, World-in-World, QuantiPhy, and BEHAVIOR-1K assess physical understanding, dynamics prediction, or closed-loop planning~\cite{chow2025physbench,xu2025deepphy,upadhyay2026worldbench,zhang2025world,puyin2025quantiphy,li2023behavior}. While valuable, they focus primarily on scenes, objects, and world dynamics rather than the semantic understanding of tangible control interfaces and the verification of action outcomes.

A second line studies agents that operate web, desktop, or mobile interfaces, including Mind2Web, WebArena, AndroidWorld, Mobile-Bench, OSWorld, ScreenSpot-Pro, and UINavBench~\cite{deng2023mind2web, zhou2024webarenarealisticwebenvironment, rawles2024androidworld, deng2024mobilebenchevaluationbenchmarkllmbased,xu2026mobilebenchv2realisticcomprehensivebenchmark, xie2024osworld, li2025screenspotproguigroundingprofessional, tan2025cradle,Agrawal_2025_ICCV_UINavBench}. More recent environment-coupled benchmarks such as HomeBench and SimuHome begin to consider device-state reasoning~\cite{li2025homebenchevaluatingllmssmart, seo2026simuhometemporalenvironmentawarebenchmark}, but they still largely rely on simulated or API-level interaction. In contrast, \projname\ focuses on tangible control interfaces that couple UI semantics with physical device states. SWITCH evaluates whether a model can interpret a real control panel, choose an action, predict the resulting state transition, and verify that the intended physical outcome has been achieved.



\section{Semantic World Interface Tasks}
SWITCH casts TCI use as interactive reasoning. While existing benchmarks (\eg PhysToolBench~\cite{zhang2025phystoolbench}) ask if a model can predict an action, SWITCH asks if it can tell whether the action has actually succeeded. Unlike in passive or open-loop settings, it links instructions to actions, consequences, and recovery. This is challenging due to hidden control logic, delayed effects, and subtle visual feedback.

To evaluate this closed-loop cycle, SWITCH introduces three progressive stages: \textit{(1) Interface Understanding} (grounding states and controls), \textit{(2) Action and Consequence Reasoning} (tracking actions and predicting temporal effects), and \textit{(3) Closed-loop Reflection} (verifying outcomes, detecting failures, and re-planning). This framing transforms isolated perception into a structured test of embodied interaction.

Questions are derived from real-world egocentric videos with dense, language-grounded annotations. Tasks span multiple-choice and open-ended formats using rubric-based LLM judging. Section~\ref{sec:benchmark_design} details the design principles, and Section~\ref{sec:data_curation} describes data curation.

\subsection{Benchmark Design}
\label{sec:benchmark_design}
SWITCH is designed to evaluate whether multimodal models can move from \textit{perceiving} an interface to \textit{acting} on it, and ultimately to \textit{checking whether the action worked}. This emphasis is critical because robust perception alone does not guarantee reliable interaction: a model may correctly read text on a panel or localize a button, yet fail to notice that the device did not enter the desired state. We therefore organize the benchmark around the three-stage closed-loop structure introduced above.

\paragraph{Interface Understanding} covers the perceptual and semantic grounding required before interaction. It includes task-aware visual question answering, where models answer goal-conditioned queries about device states, and semantic UI grounding, where they identify relevant controls and their functional roles. The objective is not generic object recognition, but grounding natural language instructions in actionable interface elements.

\paragraph{Action and Consequence Reasoning} assesses whether models can progress from the current state to the next operation and anticipate its effects. This includes action reasoning, which infers target controls, interaction modes, and fine-grained steps, and state-transition prediction, which anticipates changes in the interface, the environment, or both. This stage tests whether models understand TCI operations as causal events rather than isolated visual changes.

\paragraph{Closed-loop Reflection} evaluates whether an interaction actually succeeded. Spanning result verification, failure detection, and recovery reasoning, this stage requires models to identify visual evidence for task completion, recognize incorrect or incomplete outcomes, and select corrective actions. This phase is the cornerstone of SWITCH: passive perception is insufficient if a model cannot use environmental feedback to verify and course-correct.

\subsection{Data Curation }
\label{sec:data_curation}
SWITCH is curated through four progressive stages: real-world video collection, temporal interaction annotation, task construction, and quality-controlled splitting. The overarching goal is to preserve the complete closed-loop structure of TCI use: a user starts from a language goal, interacts with a device, observes state changes, verifies the outcome, and optionally recovers from failed or incomplete executions.

\paragraph{Video Collection.}
We collect egocentric videos of real-world interactions with everyday tangible control interfaces. We collect egocentric videos of real-world interactions with everyday tangible control interfaces. The collection comprises 1,170 egocentric video trajectories, capturing interactions with a wide array of device variants across nine distinct domains. Instead of recording isolated button presses, we instruct collectors to capture complete, task-oriented interaction segments—encompassing preparation, direct manipulation, waiting, visual inspection, and outcome checking. This design preserves the critical temporal evidence needed for action reasoning, state-transition prediction, and closed-loop verification.

\paragraph{Temporal Annotation.}
Each video is densely annotated as a sequence of interaction events linking states, actions, and outcomes. For states, annotators describe both interface-level evidence (\eg indicator lights, display text, selected modes) and environment-level evidence (\eg lighting changes, water flow, temperature shifts). For actions, we explicitly distinguish \textit{task-execution actions} (\eg pressing, rotating, swiping) from \textit{verification actions} (\eg explicitly checking system status). Crucially, we also annotate failure and recovery behaviors when an execution is incorrect, incomplete, or requires corrective follow-up.

\paragraph{Task Construction.}
From these annotations, we derive evaluation tasks perfectly aligned with our three-stage framework. \textit{Interface Understanding} evaluates task and state comprehension alongside semantic UI grounding. \textit{Action and Consequence Reasoning} tests action reasoning, sequential action quantification, and state-transition prediction. \textit{Closed-loop Reflection} rigorously evaluates outcome verification and recovery reasoning. The benchmark supports three parallel evaluation formats: multiple-choice questions (MCQs), open-ended VideoQA, and generative world-model tasks. MCQs provide scalable option-based evaluation, OpenQA evaluates free-form responses using reference answers and strict rubrics, and generative world-model tasks map an initial observation and a language action to the expected post-action video, assessing scene fidelity, control correctness, procedural correctness, and outcome observability.

\paragraph{Distractor and Hard-Negative Generation.}
To ensure robust evaluation, distractors for MCQs are carefully generated to be visually and semantically plausible rather than random. We employ source-aware and similarity-based sampling to construct hard negatives from related devices, actions, or states. In addition, we introduce Seedance-based generative distractors to synthesize fine-grained hard-negative image and video options, further increasing the diversity and difficulty of the candidate set. We stringently filter items that can be answered via language priors or spurious correlations, forcing models to genuinely rely on visual-temporal evidence rather than superficial answer-option cues.

\paragraph{Quality Control and Strategic Splitting.}
Across all construction stages, our pipeline initially yielded more than 27,000 candidate QA pairs and generative items. To guarantee an exceptionally high evaluation standard, every retained annotation and constructed question was manually inspected and reviewed, in addition to automatic validation. We removed ambiguous questions, broken media links, near-duplicates, inconsistent annotations, and items answerable without visual evidence. This stringent filtering curates a highly refined final benchmark of 8,797 visually grounded instances. Rather than competing purely on scale, SWITCH distinguishes itself through \textbf{high interaction density}—each instance is deeply embedded in a multi-stage, temporally grounded, verification-aware closed loop. To prevent data leakage, we split the benchmark strictly by source video. Finally, to focus on true reasoning bottlenecks, we derive a diagnostic subset by isolating items with near-saturated performance under a representative baseline (Qwen2.5-VL-7B-Instruct). This subset enforces the benchmark's core philosophy: to test whether a model can genuinely comprehend the implication of an outcome, rather than merely observing a visual state change without understanding underlying task success.

\begin{table*}[t]
\centering
\small
\setlength{\tabcolsep}{0pt}
\begin{tabular*}{\textwidth}{@{\extracolsep{\fill}} l ccc ccc cc c @{}}
\toprule
\multirow{2}{*}{\textbf{Model}} 
& \multicolumn{3}{c}{\textbf{Interface Understanding}} 
& \multicolumn{3}{c}{\textbf{Action \& Consequence Reasoning}} 
& \multicolumn{2}{c}{\textbf{Closed-loop Reflection}} 
& \multirow{2}{*}{\textbf{Overall}} \\
\cmidrule{2-4} \cmidrule{5-7} \cmidrule{8-9}
& \makecell{Task\\Under.} & \makecell{State\\Perc.} & \makecell{UI\\Recog.} 
& \makecell{Action\\Reas.} & \makecell{Seq.\\Quant.} & \makecell{State\\Trans.} 
& \makecell{Outcome\\Verif.} & \makecell{Recovery\\Reas.} 
& \\
\midrule
\multicolumn{10}{l}{\textit{Proprietary / API Models}} \\
\addlinespace
Gemini 3.1 Pro 
& 52.24 & \underline{37.95} & \underline{63.93} 
& \textbf{43.45} & 34.03 & \textbf{53.75} 
& \textbf{44.76} & \textbf{43.67} 
& \textbf{43.72} \\
Claude 4.7 Opus 
& \textbf{55.51} & \textbf{39.09} & 54.10 
& 38.52 & 37.25 & 49.36 
& 39.80 & 40.81 
& \underline{40.82} \\
Doubao-Seed 2.0 Pro 
& \underline{53.06} & 27.82 & 59.84 
& 36.90 & \textbf{40.20} & \underline{52.65} 
& \underline{40.71} & 40.21 
& 39.85 \\
GPT-5.5 
& 44.03 & 25.82 & \textbf{65.38} 
& \underline{39.66} & 38.38 & 50.46 
& 39.04 & \underline{43.52} 
& 39.77 \\
Grok 4.3 
& 43.67 & 20.11 & 53.28 
& 32.76 & \underline{38.66} & 47.17 
& 32.68 & 38.70 
& 34.66 \\
MIMO-V2.5 
& 41.56 & 19.12 & 56.20 
& 31.63 & 35.71 & 44.79 
& 28.98 & 29.22 
& 31.84 \\
GLM-5V Turbo 
& 42.45 & 23.68 & 40.98 
& 30.94 & 24.37 & 39.67 
& 29.68 & 29.96 
& 31.65 \\
\midrule
\multicolumn{10}{l}{\textit{Open-Weight Models}} \\
\addlinespace
Qwen3-VL-235B 
& 40.82 & 18.26 & 51.64 
& 30.59 & 32.91 & 42.78 
& 28.21 & 29.22 
& 30.66 \\
Llama-4-Maverick 
& 32.51 & 16.12 & 28.85 
& 30.39 & 14.99 & 31.44 
& 27.93 & 25.75 
& 26.25 \\
MiniMax-VL-01-Fast 
& 31.43 & 21.11 & 35.25 
& 21.58 & 23.25 & 24.31 
& 22.14 & 28.31 
& 23.39 \\
Mistral-Large-2512 
& 30.61 & 24.54 & 23.77 
& 21.18 & 23.25 & 29.98 
& 18.30 & 23.19 
& 22.49 \\
\bottomrule
\end{tabular*}
\caption{Accuracy (\%) of leading multimodal models on the strong-baseline-filtered hard subset of \projname. Items correctly answered by Qwen2.5-VL-7B-Instruct are removed. Results are organized by three closed-loop interaction capabilities and eight fine-grained skills. The best and second-best performing models in each metric are \textbf{bolded} and \underline{underlined}, respectively.}
\label{tab:main_results}
\end{table*}

\section{Experiments}

We evaluate \projname under three complementary settings: Multiple-Choice QA, OpenQA, and Interactive Video Generation. The first two share a common task formulation and multimodal context, and jointly assess closed-loop interactive reasoning under answer selection and free-form generation. The third probes generative world modeling by testing whether models can continue egocentric device-interaction trajectories while preserving control semantics, procedural structure, and verifiable outcomes. Together, these settings provide a unified view of the capabilities required for TCI-centered closed-loop interaction.


\subsection{Benchmarking Closed-Loop Interactive Reasoning with MCQ and OpenQA}
We begin with MCQ and OpenQA, focusing on three questions: (i) how well current models solve the strong-baseline-filtered multiple-choice benchmark, (ii) which interaction skills remain the most difficult, and (iii) whether open-ended evaluation exposes failures beyond option selection.

\subsubsection{Experimental Setup}
\label{sec:experimental_setup}
\paragraph{Models.}
We benchmark a comprehensive suite of frontier multimodal models, stratified into two regimes. The proprietary/API group includes Gemini 3.1 Pro~\cite{gemini31pro2026}, Claude 4.7 Opus, Doubao-Seed 2.0 Pro~\cite{seedance2026}, GPT-5.5, Grok 4.3, MIMO-V2.5, and GLM-5V Turbo. The open-weight group features Qwen3-VL-235B, Llama-4-Maverick, MiniMax-VL-01-Fast, and Mistral-Large-2512. This selection spans a diverse spectrum of parameter scales and architectural strengths across image understanding, temporal video reasoning, and instruction following.

\paragraph{Multiple-Choice Evaluation.}
For MCQ tasks, models receive the language query alongside the corresponding egocentric visual context. Notably, to rigorously evaluate \textbf{joint state reasoning}—where real-world actions trigger simultaneous physical environment and UI changes—our benchmark incorporates both standard single-choice and complex multi-select questions. For multi-select items, models must identify all applicable visual state changes, evaluated under a strict Exact Match (EM) metric without partial credit. This strict criterion reflects the reality that incomplete state perception inevitably leads to incorrect verification or recovery decisions in real-world interactive systems. Unless otherwise specified, our primary MCQ analysis is conducted on the diagnostic subset (introduced in Section \ref{sec:data_curation}). This targeted evaluation ensures that models are tested against true interactive reasoning bottlenecks, effectively mitigating the influence of language priors and superficial answer-option correlations.

\paragraph{Open-Ended Evaluation (OpenQA).}
Unlike MCQs, OpenQA requires models to generate free-form responses, testing whether they can articulate underlying states, actions, or verification logic without relying on candidate options. To ensure rigorous and reproducible scoring, we employ a \textbf{Diagnostic LLM-as-a-Judge protocol}. Rather than loose semantic matching, our judge enforces strict binary grading based on predefined rubrics. Specifically, it penalizes hallucinated visual details and requires \textit{minimal sufficient specificity} (\eg distinguishing "red" vs. "green" buttons) and \textit{temporal consistency}. Model refusals, formatting failures, and responses that exhibit severe ambiguity are strictly classified as incorrect. The detailed type-specific grading criteria and required judge output schema are provided in Appendix~\ref{app:openqa_judge}, including Table~\ref{tab:openqa_type_rules} and Table~\ref{tab:openqa_judge_schema}.
\paragraph{Metrics.}
For both MCQ and OpenQA tasks, the primary evaluation metric is strict Accuracy (\%). For a highly granular failure analysis, our diagnostic judge also yields categorized error tracking (\eg missing objects, wrong temporal order, or failed outcome implication), which we dissect in Appendix~\ref{app:case_studies}. Table~\ref{tab:main_results} reports the main multiple-choice results on the strong-baseline-filtered subset, where each model is evaluated across the eight fine-grained skills defined by our closed-loop interaction framework. To complement this option-based evaluation, Table~\ref{tab:openqa_skill_results} presents the corresponding no-options OpenQA results, which test whether models can generate the required state, action, verification, or recovery information without relying on answer choices.

\subsubsection{Capability Bottlenecks in Closed-Loop Reasoning}
\label{sec:observations}

Tables~\ref{tab:main_results} and~\ref{tab:openqa_skill_results} summarize model performance under multiple-choice and open-ended evaluations. Beyond aggregate accuracy, the skill-level breakdown reveals systematic bottlenecks in TCI-centered closed-loop interaction. Across models, failures are driven less by an inability to recognize interface elements than by difficulty connecting actions, state changes, and task completion.

\begin{table*}[t]
\centering
\small
\setlength{\tabcolsep}{0pt}
\begin{tabular*}{\textwidth}{@{\extracolsep{\fill}} l cc ccc cc c @{}}
\toprule
\multirow{2}{*}{\textbf{Model}} 
& \multicolumn{2}{c}{\textbf{Interface Understanding}} 
& \multicolumn{3}{c}{\textbf{Action \& Consequence Reasoning}} 
& \multicolumn{2}{c}{\textbf{Closed-loop Reflection}} 
& \multirow{2}{*}{\textbf{Overall}} \\
\cmidrule{2-3} \cmidrule{4-6} \cmidrule{7-8}
& \makecell{Task\\Under.} & \makecell{State\\Perc.} 
& \makecell{Action\\Reas.} & \makecell{Seq.\\Quant.} & \makecell{State\\Trans.} 
& \makecell{Outcome\\Verif.} & \makecell{Recovery\\Reas.} 
& \\
\midrule
Doubao-Seed 2.0 Pro
& 16.4 & \textbf{9.0} 
& \textbf{8.9} & \underline{33.8} & \textbf{45.0}
& 2.5 & \textbf{9.1}
& \textbf{18.9} \\
MIMO-V2.5
& 6.0 & \underline{6.0} 
& 6.4 & \textbf{36.7} & 20.0
& 3.7 & 4.5
& \underline{17.6} \\
Claude 4.7 Opus
& 3.0 & 4.3 
& \underline{7.9} & \underline{33.8} & 35.0
& 4.9 & 0.0
& 16.5 \\
GPT-5.5
& 4.5 & 3.4 
& 5.9 & 34.9 & 20.0
& 7.4 & 0.0
& 16.3 \\
GLM-5V Turbo
& 7.5 & 6.0 
& 5.4 & 32.2 & 15.0
& \textbf{9.9} & 4.5
& 16.2 \\
Grok 4.3
& 9.0 & 4.7 
& \underline{7.9} & 29.5 & 20.0
& 4.9 & \textbf{9.1}
& 15.3 \\
Llama-4-Maverick
& 3.0 & 2.6 
& 4.4 & 30.3 & 20.0
& 3.7 & 0.0
& 13.7 \\
Gemini 3.1 Pro
& 1.5 & 4.7 
& \underline{7.9} & 25.5 & 20.0
& 6.2 & 0.0
& 13.2 \\
Mistral-Large-2512
& 1.5 & 2.6 
& 6.9 & 26.0 & 20.0
& 6.2 & 0.0
& 12.7 \\
MiniMax-VL-01-Fast
& 7.5 & 2.6 
& 6.4 & 20.9 & \textbf{45.0}
& \underline{8.6} & 0.0
& 11.8 \\
Qwen3-VL-235B
& 4.5 & 3.8 
& 4.9 & 22.3 & 40.0
& 1.2 & \textbf{9.1}
& 11.6 \\
\bottomrule
\end{tabular*}
\caption{OpenQA accuracy (\%) on the 1000 no-options subset of \projname. Responses are evaluated by a strict rubric-based binary LLM judge conditioned on the question, gold answer, and task-specific rubric. Note that UI Recognition is excluded as this specific OpenQA subset contains no spatial-grounding UI items. }
\label{tab:openqa_skill_results}
\end{table*}

\paragraph{1/ Closed-loop interaction remains far from solved.}
On the diagnostic MCQ subset, the strongest model reaches only 43.72\% overall accuracy, and the best open-weight model reaches 30.66\%. Since this subset already removes items solved by a strong open-source VLM baseline, this indicates that the remaining examples require visual-temporal and feedback-based reasoning beyond common language priors.

\paragraph{2/ UI recognition outpaces state perception.}
A clear asymmetry appears within Interface Understanding. GPT-5.5 achieves 65.38\% on UI Recognition but only 25.82\% on State Perception, a 39.56-point gap. This pattern holds across models: even the best State Perception score is below 40\%. Thus, recognizing where a control is does not imply understanding what state the device is in, especially when the relevant evidence involves selected modes, numeric values, indicator lights, or subtle environmental feedback.

\paragraph{3/ Temporal action tracking is brittle.}
Sequential Action Quantification remains difficult across models: the best MCQ score is only 40.20\%, and the best OpenQA score is 36.7\%. These results suggest that models struggle to maintain temporal consistency over longer video segments, especially when distinguishing task-execution steps from procedural or verification behaviors.

\paragraph{4/ Verification and recovery expose the closed-loop gap.}
Outcome Verification and Recovery Reasoning remain weak even for the best models, with top MCQ scores of 44.76\% and 43.67\%, respectively. Under OpenQA, both abilities drop below 10\%. This indicates that models struggle not only to select actions, but also to determine whether an action achieved the intended goal and what corrective replanning is required after failure.

\paragraph{5/ OpenQA reveals reliance on answer-option scaffolding.}
The best MCQ overall accuracy is 43.72\%, whereas the best OpenQA accuracy on the no-options \textbf{hard1000} subset is only 18.9\%. For GPT-5.5, accuracy drops from 39.77\% in MCQ to 16.3\% in OpenQA. This gap suggests that multiple-choice options provide substantial semantic scaffolding; without choices, models must directly generate the relevant state, action, count, or verification logic, leading to many partial, vague, or contradictory responses under rubric-based judging.To further unpack these failures, Appendix~\ref{app:openqa_diagnostics} reports a diagnostic breakdown of OpenQA responses in Table~\ref{tab:openqa_diagnostics}, separating fully correct answers from partial matches, contradictions, and complete misses.

\subsection{Probing Generative World Modeling through Interactive Video Generation}
\label{subsec:probing_world_modeling}

\projname probes whether current generative models exhibit usable world modeling when asked to continue egocentric device-interaction scenes under explicit task constraints. The goal is not to assess aesthetics, but to test whether models can preserve input scene, operate the correct control, follow the intended action sequence, and make outcomes visually verifiable. We evaluate both image-conditioned and video-conditioned generation.

\paragraph{Experiment Setup}
As shown in Fig~\ref{fig:firstPic}, \projname covers nine categories. For each category, we sample 20 image-conditioned instances and 5 video-conditioned instances, yielding 180 and 45 inputs, respectively. In the image-conditioned setting, each model receives one egocentric anchor frame and a prompt specifying the goal, required action sequence, stop condition, visible evidence, and invariance constraints. We test six models: five proprietary systems, \texttt{Doubao-Seed 2.0}~\cite{seedance2026}, \texttt{Kling3.0}~\cite{klingteam2025klingomnitechnicalreport}, \texttt{Runway Gen-4.5}\footnote{https://runwayml.com/research/introducing-runway-gen-4.5}, \texttt{Veo3.1}~\cite{veo3}, and \texttt{Wan2.7}~\cite{wan2025wanopenadvancedlargescale}, and one open-source model, \texttt{Wan2.2}~\cite{wan2025wanopenadvancedlargescale}. In the video-conditioned setting, each model receives an egocentric clip and a continuation prompt; we evaluate \texttt{Doubao-Seed 2.0} and \texttt{Wan2.7}. In total, we evaluate 1,170 generated videos: 1,080 in the image-conditioned setting and 90 in the video-conditioned setting. Detailed generation parameters are given in Appendix~\ref{app:generation_parameters}.

\paragraph{Prompting and evaluation.}
Prompts explicitly constrain task type, next action, stop condition, required visible evidence, viewpoint, device layout, and object identity. Image-conditioned generation, therefore, requires the model to infer a plausible future interaction from a single frame while preserving the original scene structure. Video-conditioned generation requires the model to continue the scene \emph{after} the observed prefix, rather than replaying or rewriting it. These settings directly test whether the model can maintain latent state, control-function binding, and causal consequences over time.

All generated videos are evaluated by \texttt{Gemini 3.1 Pro}~\cite{gemini31pro2026} in a VLM-as-Judge setup. The judge is instructed to rely only on visible evidence and not to reward mere visual plausibility. Each video receives 1--5 scores on four dimensions: \textit{scene\_fidelity}, \textit{control\_correctness}, \textit{procedure\_correctness}, and \textit{execution\_observability}. The judge also outputs an overall score, a confidence score, and major error descriptions. The full evaluation prompt is provided in Appendix~\ref{app:vlm_judge_prompt}.

\begin{table*}[t]
\centering
\small
\setlength{\tabcolsep}{4pt} 
\resizebox{\linewidth}{!}{
\begin{tabular}{llcccccccccc}
\toprule
& & \multicolumn{5}{c}{\textbf{VLM as Judge}} & \multicolumn{5}{c}{\textbf{HumanEval}} \\
\cmidrule(lr){3-7} \cmidrule(lr){8-12}
\textbf{Input Form} & \textbf{Model} & {\makecell[c]{\textbf{Scene}\\ \textbf{Fidelity}}}& {\makecell[c]{\textbf{Control}\\ \textbf{Correctness}}} & {\makecell[c]{\textbf{Procedure}\\\textbf{Correctness}}}  & {\makecell[c]{\textbf{Execution} \\ \textbf{Observability}}}& \textbf{Overall} & {\makecell[c]{\textbf{Scene}\\ \textbf{Fidelity}}}& {\makecell[c]{\textbf{Control}\\ \textbf{Correctness}}} & {\makecell[c]{\textbf{Procedure}\\\textbf{Correctness}}}  & {\makecell[c]{\textbf{Execution} \\ \textbf{Observability}}}& \textbf{Overall} \\
\midrule
\multirow{7}{*}{\makecell[c]{Single Image\\to Video}} 
& Doubao-Seed 2.0 & \textbf{2.96} & \textbf{2.96} & \textbf{3.04} & \textbf{4.39} & \textbf{2.64} & \textbf{4.05} & 3.28 & 3.77 & \textbf{4.65} & 3.52 \\
& Kling 3.0 & 1.86 & 2.36 & 2.42 & 3.83 & 1.87 & 3.39 & 3.53 & 3.65 & 4.02 & 3.37 \\
& Runway Gen-4.5 & 1.61 & 2.25 & 2.38 & 3.56 & 1.80 & 3.87 & \textbf{4.09} & \textbf{4.83} & 4.37 & 3.44 \\
& Veo 3.1 & 2.19 & 2.47 & 2.70 & 4.19 & 2.09 & 3.55 & 3.59 & 4.27 & 4.62 & \textbf{3.79} \\
& Wan 2.7 & 2.04 & 1.98 & 2.02 & 3.60 & 1.72 & 3.45 & 3.37 & 3.62 & 3.73 & 3.02 \\
& Wan 2.2 & 1.55 & 1.57 & 1.49 & 2.99 & 1.38 & 2.49 & 2.97 & 3.20 & 4.27 & 2.87 \\
& \textbf{Average} & 2.04 & 2.26 & 2.34 & 3.76 & 1.92 & 3.47 & 3.47 & 3.89 & 4.28 & 3.33 \\
\midrule
\multirow{3}{*}{\makecell[c]{Video\\to Video}} 
& Doubao-Seed 2.0  & \textbf{2.90} & \textbf{3.07} & \textbf{3.13} & \textbf{4.39} & \textbf{2.54} & \textbf{3.98} & \textbf{3.68} & \textbf{3.95} & \textbf{4.40} & \textbf{3.65} \\
& Wan 2.7 & 2.49 & 2.37 & 2.10 & 3.81 & 1.83 & 3.67 & 2.68 & 3.05 & 3.43 & 2.86 \\
& \textbf{Average} & 2.70 & 2.72 & 2.62 & 4.10 & 2.18 & 3.83 & 3.18 & 3.50 & 3.92 & 3.25 \\
\bottomrule
\end{tabular}
}
\caption{\textbf{VLM-as-Judge and HumanEval results under different input forms.} The first five columns present the VLM-as-Judge results computed from 1,170 valid evaluations. The last five columns show the HumanEval results computed from 160 human-evaluated videos. Single-image input refers to generating a video from an initial frame and a text prompt, while video input denotes generating a video conditioned on a short video clip and a text prompt. Each model row reports the mean score over its evaluated samples, and each Average row reports the arithmetic mean of the listed models within the same input form. All values are averaged on a 1--5 scale, and the best results in each section are highlighted in \textbf{bold}.}
\label{tab:generate_combined_results}
\end{table*}

\begin{figure*}[htbp]
\centering
\includegraphics[width=1\linewidth]{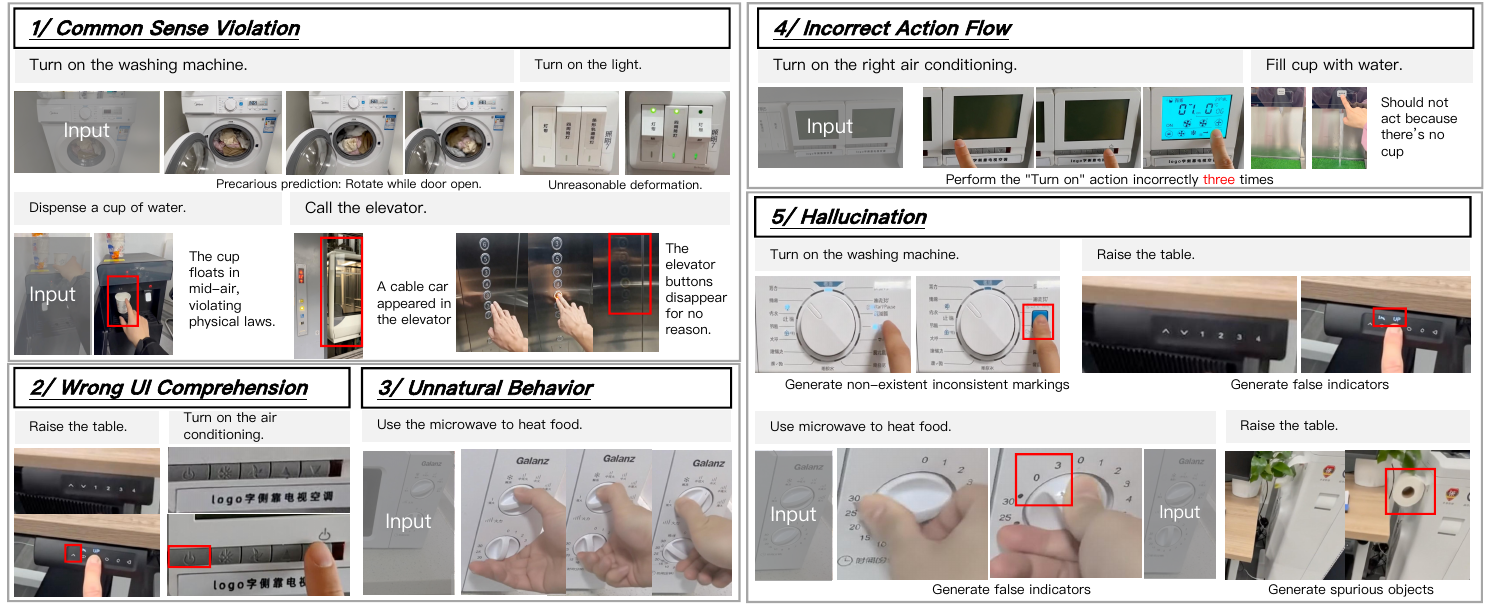}
\caption{Case studies of representative errors in video generation. Cropped image frames to focus on issues.}

\label{fig:veo3cases}
\end{figure*}

\paragraph{Automatic results.}
The VLM-as-Judge results in Table~\ref{tab:generate_combined_results} point to a clear gap between visual plausibility and task grounding. Across both input forms, models are much better at making an interaction look legible than at making it correct: \textit{execution\_observability} is consistently the strongest dimension (3.76 for image-to-video and 4.10 for video-to-video on average), whereas \textit{scene\_fidelity}, \textit{control\_correctness}, and \textit{procedure\_correctness} remain substantially weaker. In practice, many generations show a recognizable hand--device interaction, but the hand often targets the wrong control, drifts from the original scene, or skips the required action order. Video conditioning brings only a modest improvement over single-image prompting (overall 2.18 vs.\ 1.92), suggesting that observing a short prefix helps continuity, but does not solve the harder problem of tracking device state and causal task structure. \texttt{Doubao-Seed 2.0} performs best in both settings, but the gain is incremental rather than qualitative.

\paragraph{Human evaluation.}
Human evaluation on 160 videos from four categories, each rated by three blind annotators, leads to the same substantive conclusion. As shown in Table~\ref{tab:generate_combined_results}, although human raters assign higher absolute scores than the automatic judge and slightly reshuffle the image-conditioned model ranking, they identify the same bottleneck: current models are usually better at preserving appearance and rendering visible interactions than at binding the correct control to the intended goal and executing the steps in the right order. Taken together, the automatic and human results suggest that the main failure is not a lack of realistic motion, but a lack of reliable latent state tracking and action semantics.

\paragraph{Case study.}
Figure~\ref{fig:veo3cases} presents representative failures. These examples expose five recurring failure modes. First, the model violates commonsense or physical constraints, such as rotating a washing machine drum while the door is open, deforming a light switch, or producing impossible elevator states. Second, it misidentifies the relevant UI region or physical control, pressing the wrong button or panel area. Third, it produces unnatural motor behavior, such as attempting to manipulate multiple microwave knobs simultaneously with one hand. Fourth, it breaks procedural logic by repeating actions, ignoring necessary preconditions, or producing outcomes unsupported by the scene, such as dispensing water without a cup. Fifth, it hallucinates UI elements or objects, including nonexistent markings, false indicators, and spurious scene objects. These failures are often visually convincing at a glance, but they reveal weak latent modeling of device state, action preconditions, and causal consequences.

\section{Conclusion}
We present \projname, a benchmark for evaluating closed-loop interactive reasoning with tangible control interfaces in real-world egocentric environments. SWITCH organizes evaluation into \textbf{8 tasks} spanning \textbf{three core capabilities}: \textit{Interface Understanding}, \textit{Action \& Consequence Reasoning}, and \textit{Closed-loop Reflection}. By moving beyond static perception and open-loop action prediction, SWITCH measures whether models can interpret interface semantics, reason about actions and their consequences, and verify task success from visual feedback in realistic control settings.

Experiments on VLMs and generative models highlight substantial room for improvement. For VLMs, performance remains limited on fine-grained interface grounding, action-and-consequence reasoning, and outcome verification, indicating that general multimodal ability does not yet translate into reliable real-world control understanding. For generative models, both VLM-as-Judge and human evaluation show that visually plausible interaction videos often fail to preserve the correct scene, manipulate the right control, or follow the intended procedure. Taken together, these results position \projname as a challenging and practical testbed for advancing grounded multimodal reasoning, closed-loop verification, and world modeling for real-world interactive environments. We believe \projname can serve as a strong foundation for developing and evaluating multimodal systems for reliable real-world interaction.

\section*{Limitations}

This work has several limitations. First, although \projname covers over 1,170 device interactive videos across 9 domains, it does not yet capture the full diversity of tangible control interfaces encountered in everyday life. The current release is still concentrated in a limited set of environments and interface styles, and may underrepresent region-specific, language-specific, and industrial control settings. In future iterations, we plan to expand the data collection to more diverse and more international scenarios, with broader geographic, linguistic, and device coverage.

Second, SWITCH is currently centered on offline evaluation from recorded egocentric videos, question answering, and generative continuation. While this setup is useful for controlled analysis, it does not fully measure the challenges of real-time embodied interaction, such as actuation errors, latency, safety constraints, and error accumulation during execution. In future versions, we plan to incorporate more real-machine interaction data and evaluation settings that better reflect online decision making and physical execution.

Third, some evaluation dimensions are less well covered than others. In particular, verification- and recovery-related cases are harder to collect because they often require delayed effects, failed attempts, or multi-step correction trajectories. As a result, these cases are relatively less abundant than standard task-execution examples in the current release. To address this, we plan to collect more failure-driven, delayed-outcome, and multi-step recovery data in order to strengthen the benchmark's coverage of closed-loop reflection.

Finally, the generative world-model evaluation remains an approximation of interactive competence. The automatic protocol relies on a VLM-as-Judge setup, and human evaluation is conducted on a subset of categories and models rather than the full benchmark. Although these protocols provide useful signals, they may not capture all subtle errors in causal correctness or long-horizon procedural consistency. In future work, we plan to expand human evaluation coverage, improve the automatic judging pipeline, and explore more direct evaluation protocols for generated interaction trajectories.

\section*{Acknowledgments}
We gratefully acknowledge the valuable support provided by Zhongyuan Wang throughout this work. We also sincerely thank Runze Xiao, Aoyang Cai, and Ruochuan Shi for their insightful revision comments.


\bibliography{main_v2}

\newpage
\appendix

\section{Evaluation Details}
\label{app:evaluation_details}

\subsection{MCQ Evaluation}
\label{app:mcq_eval}

For multiple-choice evaluation, each item contains a multimodal query, one or more visual inputs, and a fixed set of candidate answers. Depending on the task format, answer candidates may be text, images, videos, or numeric values. Models are prompted to return only the selected option label or labels. We parse the model response into option labels and compare it with the ground-truth answer.

\paragraph{Single-choice questions.}
For standard MCQ items, a prediction is counted as correct if the parsed option label exactly matches the ground-truth option. Invalid outputs, empty responses, or responses that cannot be mapped to a valid option are counted as errors and receive zero credit.

\paragraph{Multi-select questions.}
A subset of outcome-verification questions is multi-select. These items ask models to select all evidence options that demonstrate task success, such as all images or states showing that an operation worked. For these questions, the prediction is counted as correct only if the predicted option set exactly matches the ground-truth set. We additionally compute partial scores for analysis by comparing selected correct and incorrect evidence options, but the main accuracy metric uses exact set matching.

\paragraph{Strong-baseline-filtered hard subset.}
To construct a harder diagnostic subset, we remove validation and test items correctly answered by Qwen2.5-VL-7B-Instruct. This filtering reduces the influence of language priors and answer-option shortcuts, and focuses evaluation on examples not solved by a strong open-source VLM baseline. All main MCQ results are reported on this filtered subset unless otherwise specified.

\paragraph{Skill aggregation.}
We aggregate item-level accuracy into the eight evaluation skills used in the main results. Task Understanding uses \texttt{vqa\_task}; State Perception uses \texttt{vqa\_state}; UI Recognition uses \texttt{ui\_grounding}; Action Reasoning uses action questions excluding counting; Sequential Action Quantification uses action-counting questions; State Transition Prediction uses \texttt{final\_state}; Outcome Verification uses \texttt{verification\_state} and \texttt{verification\_action}; Recovery Reasoning uses \texttt{recovery}. Overall accuracy is computed over all evaluated items.


\subsection{OpenQA Rubric-Based Judge}
\label{app:openqa_judge}

For OpenQA, we use a strict rubric-based binary judge. The judge receives the question, gold answer, optional task-specific rubric, and model prediction, and returns a binary correctness decision with diagnostic labels. The judge is instructed to grade semantic correctness rather than surface similarity. Paraphrases, translated device terms, and equivalent UI terminology are accepted when they refer to the same object, state, action, or result. However, missing essential attributes or contradictions make the answer incorrect. Partial answers are recorded diagnostically but receive zero credit.

\paragraph{Contamination check.}
Before grading, the judge checks whether an OpenQA item appears to be a multiple-choice question mistakenly included in the OpenQA set. Items referring to options, four choices, A/B/C/D, or bare option-letter gold answers are flagged as \texttt{multiple\_choice\_contamination}. Such items receive a score of 0 if the option contents are unavailable.

\paragraph{Type-specific grading rules.}
Table~\ref{tab:openqa_type_rules} summarizes the strictness criteria used by the judge for different question types.

\begin{table}[t]
\centering
\small
\setlength{\tabcolsep}{3pt}
\begin{tabular}{lp{0.64\columnwidth}}
\toprule
\textbf{Type} & \textbf{Correctness criterion} \\
\midrule
Counting & Exact numeric answer is required unless the rubric allows tolerance. \\
Yes/No & The answer must clearly affirm or negate the proposition. \\
Grounding & The object, region, UI element, state, or action must be specific enough to distinguish it from alternatives. \\
Action / Procedural & The correct action and target must be named; missing an essential target, direction, button, device, or attribute is incorrect. \\
Temporal & Events or actions must appear in the correct order. \\
Verification & The answer must provide the requested outcome or success/failure implication required by the question or rubric. \\
Recovery & The answer must identify the correct recovery action or correction target when recovery is requested. \\
State & Essential attributes such as color, number, text, direction, location, on/off state, mode, or selected option must be correct. \\
Other & The gold answer and rubric are applied directly. \\
\bottomrule
\end{tabular}
\caption{Question-type-specific correctness rules used by the OpenQA binary judge.}
\label{tab:openqa_type_rules}
\end{table}

\paragraph{Output schema.}
The judge returns exactly one JSON object. Table~\ref{tab:openqa_judge_schema} lists the required fields.

\begin{table}[t]
\centering
\small
\setlength{\tabcolsep}{4pt}
\begin{tabular}{lp{0.62\columnwidth}}
\toprule
\textbf{Field} & \textbf{Description} \\
\midrule
\texttt{question\_type} & Inferred task type, such as counting, verification, recovery, temporal, grounding, yes/no, action, procedural, state, or other. \\
\texttt{verdict} & Binary judgment: correct or incorrect. \\
\texttt{score} & Binary score, either 1 or 0. \\

\texttt{answer\_match} & Diagnostic label indicating exact match, paraphrase, partial match, contradiction, or complete miss. \\
\texttt{essential\_missing} & List of essential missing details, if any. \\
\texttt{wrong\_extra\_info} & Whether the prediction adds contradictory information. \\
\texttt{too\_vague} & Whether the answer is too vague to uniquely identify the required answer. \\
\texttt{format\_issue} & Whether the answer has a formatting or contamination issue. \\
\texttt{brief\_reason} & One concise sentence justifying the score. \\
\bottomrule
\end{tabular}
\caption{Required output fields for the OpenQA rubric-based binary judge.}
\label{tab:openqa_judge_schema}
\end{table}
\subsection{OpenQA Diagnostics}
\label{app:openqa_diagnostics}

\begin{table*}[t]
\centering
\small
\setlength{\tabcolsep}{0pt}
\begin{tabular*}{\textwidth}{@{\extracolsep{\fill}} l c cc ccc @{}}
\toprule
\multirow{2.5}{*}{\textbf{Model}} 
& \multirow{2.5}{*}{\textbf{\makecell{Overall \\ Correct (\%) $\uparrow$}}} 
& \multicolumn{2}{c}{\textbf{Credited (\%)} $\uparrow$} 
& \multicolumn{3}{c}{\textbf{Zero-Credit Errors (\%)} $\downarrow$} \\
\cmidrule{3-4} \cmidrule{5-7}
& & Exact & Paraphrase 
& Partial & Contradiction & Complete Miss \\
\midrule
\multicolumn{7}{l}{\textit{Proprietary / API Models}} \\
\addlinespace
Doubao-Seed 2.0 Pro 
& \textbf{18.9} & 11.8 & \textbf{7.1} 
& 21.5 & 50.7 & 8.9 \\
MIMO-V2.5 
& \underline{17.6} & \textbf{13.1} & 4.5 
& 17.5 & 59.1 & \textbf{5.8} \\
Claude 4.7 Opus 
& 16.5 & 12.6 & 3.9 
& 25.9 & 47.0 & 10.6 \\
GPT-5.5 
& 16.3 & \underline{13.0} & 3.3 
& 29.0 & 45.3 & 9.4 \\
GLM-5V Turbo 
& 16.2 & 11.9 & 4.3 
& 26.7 & 48.0 & 9.1 \\
Grok 4.3 
& 15.3 & 10.5 & \underline{4.8} 
& 24.2 & 52.8 & \underline{7.7} \\
Gemini 3.1 Pro 
& 13.2 & 9.6 & 3.6 
& 25.7 & 44.9 & 16.2 \\
\midrule
\multicolumn{7}{l}{\textit{Open-Weight \& Publicly Available Models}} \\
\addlinespace
Llama-4-Maverick 
& 13.7 & 11.2 & 2.5 
& 27.4 & \textbf{41.6} & 17.3 \\
Mistral-Large-2512 
& 12.7 & 9.7 & 3.0 
& \textbf{15.8} & 60.4 & 11.1 \\
MiniMax-VL-01-Fast 
& 11.8 & 7.8 & 4.0 
& 18.5 & 56.2 & 13.5 \\
Qwen3-VL-235B 
& 11.6 & 8.4 & 3.2 
& 22.7 & 55.9 & 9.8 \\
\bottomrule
\end{tabular*}
\caption{Diagnostic error distribution (\%) on the \textbf{\projname} OpenQA subset (no-options). Performance is evaluated using our strict rubric-based judgment. \textit{Overall Correct} serves as the primary metric, comprising the sum of \textit{Exact} and \textit{Paraphrase} matches. The remaining five columns represent the complete distribution of the model's responses (summing to 100\%). Answers flagged as \textit{Partial} overlap with the gold reference but omit essential specificity constraints, thus receiving strictly zero credit. The best performing model for each metric is \textbf{bolded}, and the second-best is \underline{underlined}.}
\label{tab:openqa_diagnostics}
\end{table*}

Table~\ref{tab:openqa_diagnostics} shows that low OpenQA accuracy is not simply an artifact of strict surface matching. The judge accepts paraphrases as correct, but many model outputs are partial or contradictory. For example, GPT-5.5 produces 290 partial answers but only 163 correct answers, suggesting that it often identifies part of the relevant interaction while missing essential details such as numeric values, action targets, success conditions, or recovery steps.


\subsection{Generation Parameters}
\label{app:generation_parameters}

Table~\ref{tab:generation_parameters} summarizes the generation settings used for each model in our experiments.



\begin{table}[t]
\centering
\small
\setlength{\tabcolsep}{8pt}
\begin{tabular}{lccc}
\toprule
\textbf{Model} & \textbf{Duration (s)} & \textbf{FPS} \\
\midrule
Seedance 2.0 & 10 & 24 \\
Kling v3 & 10 & 24 \\
Veo 3.1 & 8 & 24 \\
WAN 2.2 & 5 & 16 \\
WAN 2.7 & 10 & 30 \\
Runway Gen4.5 & 10 & 24 \\
\bottomrule
\end{tabular}
\caption{\textbf{Generation parameters used in our experiments.}}
\label{tab:generation_parameters}
\end{table}

\subsection{VLM-as-Judge Prompt}
\label{app:vlm_judge_prompt}

For each generated video, we prompt \texttt{Gemini 3.1 Pro} with the video and its corresponding task specification, including the task type, goal, next action, stop condition, required visible evidence, and scene invariance constraints. The judge is instructed to rely only on visible evidence in the generated video, rather than inferring hidden steps or unseen state changes. It is further instructed to evaluate task-grounded correctness instead of visual beauty, and to output a single JSON object. Table~\ref{tab:vlm_judge_dimensions_appendix} summarizes the four evaluation dimensions, while Table~\ref{tab:vlm_judge_schema} lists the required output fields.

\begin{table}[t]
\centering
\small

\setlength{\tabcolsep}{6pt} 

\begin{tabular}{lp{0.68\columnwidth}}
\toprule
\textbf{Dimension} & \textbf{Definition} \\
\midrule

\makecell[c]{scene\\fidelity} & Whether the visible scene remains trustworthy, physically plausible, and faithful to the input without hallucinated UI, scene redesign, viewpoint jumps, object inconsistencies, or physics violations. \\
\makecell[c]{control\\correctness} & Whether the correct control or region is operated and whether the local feedback matches the intended function. \\
\makecell[c]{procedure\\correctness} & Whether the action sequence is complete, logically correct, properly ordered, and satisfies the required stop condition. \\
\makecell[c]{execution\\observability} & Whether the key action is clearly, completely, and confirmably executed from visible evidence. \\
\bottomrule
\end{tabular}
\caption{\textbf{Core evaluation dimensions in the VLM-as-Judge protocol.}}
\label{tab:vlm_judge_dimensions_appendix}
\end{table}

\paragraph{Scoring prompt.}
The judge is instructed with the following rules:
\begin{enumerate}
    \item \textbf{General instruction.} You are an expert judge for egocentric device-interaction videos. Your task is to score whether the generated video accurately and verifiably follows real-world interaction patterns. Judge only from visible evidence in the generated video. Do not infer hidden steps, unseen states, or unshown corrections. Do not judge visual beauty; judge task-grounded correctness only.
    \item \textbf{Score scale.} Use a 1--5 scale, where 5 means excellent or strongly correct, 4 means mostly correct, 3 means partially correct, 2 means mostly incorrect, and 1 means severely incorrect.
    \item \textbf{Dimensions.} Evaluate exactly four dimensions:
    \begin{itemize}
    \item \texttt{scene\_fidelity}
    \item \texttt{control\_correctness}
    \item \texttt{procedure\_correctness}
    \item \texttt{execution\_observability}
    \end{itemize}
    \item \textbf{Dimension-specific criteria.}
    \begin{itemize}
        \item \texttt{scene\_fidelity}: measures whether the visible scene evidence itself is trustworthy. This dimension combines physical plausibility and hallucination control. It should be scored lower when the video shows fabricated UI elements, labels, buttons, indicators, or text; extra objects or rewritten device structure; viewpoint jumps or scene redesign; impossible motion, teleportation, disappearance, physics violations, broken scene invariants, or forbidden changes. It should not be used for wrong control choice, wrong step order, or incomplete but visible motion.
        \item \texttt{control\_correctness}: measures whether the correct control or region is operated and whether the local feedback matches the intended function. It should be scored lower when the video presses the wrong button, turns the wrong knob, clicks the wrong region, binds the wrong control to the wrong function, or shows local feedback inconsistent with the intended control. It should not be used for step order, missing prerequisites, or action visibility.
        \item \texttt{procedure\_correctness}: measures whether the task procedure is complete and logically correct. This dimension combines logical reasoning and action completeness. It should be scored lower when the video skips steps, violates step order, omits prerequisites, stops prematurely, includes unnecessary repetition that breaks the procedure, performs recovery without visible correction, provides insufficient verification evidence, or fails to satisfy the required stop condition. It should not be used for wrong control choice, unclear motion, or fabricated scene content.
        \item \texttt{execution\_observability}: measures whether the key action is clearly, completely, and confirmably executed. It should be scored lower when the video shows only partial execution, unclear hand-control contact, action too small to confirm, heavy occlusion, blur, truncation, or repeated ineffective motion. It should not be used for wrong control choice, step order, or fabricated UI.
    \end{itemize}
    \item \textbf{Orthogonality rules.} If the problem is fake scene content, reduce \texttt{scene\_fidelity}. If the problem is wrong control selection, reduce \texttt{control\_correctness}. If the problem is missing steps, wrong order, or invalid recovery, reduce \texttt{procedure\_correctness}. If the problem is unclear or incomplete action visibility, reduce \texttt{execution\_observability}. If a single issue affects multiple dimensions, multiple dimensions may be reduced, but each reason must describe a distinct core issue.
    \item \textbf{Task-type emphasis.} For \texttt{state\_transition\_video}, emphasize \texttt{next\_action}, \texttt{required\_evidence\_ui}, \texttt{required\_evidence\_physical}, and \texttt{stop\_condition}. For \texttt{verification\_state\_video}, emphasize whether the video shows sufficient evidence for verification. For \texttt{recovery\_video}, require visible correction first, then return to \texttt{post\_fix\_state}, and finally progression toward success.
    \item \textbf{Scoring anchors.}
    \begin{itemize}
        \item \texttt{scene\_fidelity}: 5 means stable and physically plausible; 4 means mostly stable with minor issues; 3 means visible inconsistencies but still judgeable; 2 means clear physical or scene integrity problems; 1 means severe fabrication or invariant violation.
        \item \texttt{control\_correctness}: 5 means correct control and correct feedback; 4 means mostly correct with minor ambiguity; 3 means partially correct but with weak binding or incomplete feedback; 2 means wrong control or misleading feedback; 1 means no valid control interaction or fabricated control behavior.
        \item \texttt{procedure\_correctness}: 5 means correct steps, order, prerequisites, stopping point, and verification or recovery logic; 4 means mostly correct with minor missing steps or ambiguity; 3 means partially correct with weak logical links or missing required steps; 2 means wrong order, missing prerequisites, missing key steps, or insufficient verification; 1 means the task logic is largely broken.
        \item \texttt{execution\_observability}: 5 means the key action is complete, clear, natural, and easy to confirm; 4 means mostly complete with minor stiffness or redundancy; 3 means visible but only partially executed or not fully convincing; 2 means ambiguous, incomplete, or heavily obscured; 1 means the key action is missing or cannot be confirmed.
    \end{itemize}
    \item \textbf{Overall score.} Score the overall result on the same 1--5 scale. The overall score should be consistent with the four dimension scores, but should not be computed by a fixed weighted formula; instead, it should reflect direct judgment based on the visible evidence.
    \item \textbf{Confidence.} Confidence is also rated on a 1--5 scale, where 5 means the evidence is very clear and the judgment is highly reliable, 4 means the evidence is clear with minor ambiguity, 3 means the evidence is usable but somewhat incomplete or uncertain, 2 means the evidence is weak, blurry, or partially missing, and 1 means the evidence is severely insufficient for stable judgment.
    \item \textbf{Output format.} Output only a single valid JSON object. Each dimension must contain exactly two keys, \texttt{score} and \texttt{reason}. Scores must be integers from 1 to 5. Reasons must be short, specific, and evidence-based. The output must additionally include \texttt{overall\_score}, \texttt{pass\_label}, \texttt{confidence}, and \texttt{major\_errors}. No markdown, code fences, or extra text are allowed.
\end{enumerate}

\begin{table}[t]
\centering
\small
\setlength{\tabcolsep}{5pt}
\resizebox{\linewidth}{!}{
\begin{tabular}{lp{0.56\columnwidth}}
\toprule
\textbf{Field} & \textbf{Description} \\
\midrule
\texttt{scene\_fidelity.score} & Integer score from 1 to 5. \\
\texttt{scene\_fidelity.reason} & Short evidence-based explanation. \\
\texttt{control\_correctness.score} & Integer score from 1 to 5. \\
\texttt{control\_correctness.reason} & Short evidence-based explanation. \\
\texttt{procedure\_correctness.score} & Integer score from 1 to 5. \\
\texttt{procedure\_correctness.reason} & Short evidence-based explanation. \\
\texttt{execution\_observability.score} & Integer score from 1 to 5. \\
\texttt{execution\_observability.reason} & Short evidence-based explanation. \\
\texttt{overall\_score} & Overall score on the 1--5 scale. \\
\texttt{confidence} & Confidence score from 1 to 5. \\
\texttt{major\_errors} & List of major error descriptions. \\
\bottomrule
\end{tabular}
}
\caption{\textbf{Required JSON output fields for the VLM judge.}}
\label{tab:vlm_judge_schema}
\end{table}

\begin{table*}[t]
\centering
\small
\renewcommand{\arraystretch}{1.2} 
\begin{tabularx}{\textwidth}{@{} >{\bfseries}l X @{}}
\toprule
Scene Category & \textbf{Representative Devices / Interfaces} \\
\midrule
Lighting &
Basic light-switch panels; touch lighting panels; scene-mode lighting panels; dimming and color-temperature controls; meeting-room lighting panels; bathroom light-fan combo panels. \\
\addlinespace
Sanitaryware &
Bathroom heater panels; ventilation, drying, heating, and lighting combo panels; water-heater panels; smart toilet side panels; bidet and flush controls; towel-rack or bathroom appliance panels. \\
\addlinespace
Beverage &
Water dispensers and purifiers; hot-and-cold water stations; drinking-water systems; coffee machines; beverage machines; kettles and hot-water appliances. \\
\addlinespace
Cleaning &
Washing machines; dryers; washer-dryer combo machines; dishwashers; fixed cleaning-device panels; robot-vacuum or housekeeping-device controls. \\
\addlinespace
Building &
Elevator hall-call panels; elevator cabin floor panels; public-service kiosks; queue-ticket terminals; access or public-facility control panels. \\
\addlinespace
Ecosytem &
Home air-conditioner panels; central HVAC wall controllers; thermostat and floor-heating panels; fans; air purifiers; humidifiers; dehumidifiers; environment and climate-control interfaces. \\
\addlinespace
Kitchen &
Microwaves; ovens; air fryers; stovetops; induction cookers; electric pots; rice cookers; range hoods; thermal cooking appliances; kitchen heating and processing controls. \\
\addlinespace
Furniture &
Sit-stand desks; lifting platforms; robotic lifting fixtures; motorized furniture controls; electromechanical furniture panels. \\
\addlinespace
Office &
Printers; copiers; all-in-one office machines; meeting-room control panels; information terminals; self-service terminals; PC power switches; power-strip controls. \\
\bottomrule
\end{tabularx}
\caption{Nine scene categories and representative TCI device/interface types covered by \projname. The categories strictly align with the scene taxonomy presented in the benchmark overview, encompassing a diverse spectrum of real-world interactive systems.}
\label{tab:dataset_scenarios}
\end{table*}

\begin{table}[t]
\centering
\small
\setlength{\tabcolsep}{1mm}
\resizebox{0.95\columnwidth}{!}{
\begin{tabular}{lll}
\toprule
\textbf{Category} & \textbf{Statistic} & \textbf{Number} \\
\midrule
\multirow{3}{*}{Source Coverage}
& Source dataset slugs & 24 \\
& Scenario families & 9 \\
& Interactive videos & 1,170\\
\midrule
\multirow{5}{*}{QA Construction}
& MCQ / visual QA instances & 20,075 \\
& Unique OpenQA items & 7,471 \\
& Candidate QA items total & 27,546 \\
& Hard OpenQA evaluation subset & 1,000 \\
& Multi-select MCQ items & 2,601 \\
\midrule
\multirow{8}{*}{Evaluation Skills}
& Task Understanding & 1,266 \\
& State Perception & 3,466 \\
& UI Recognition & 577 \\
& Action Reasoning & 9,567 \\
& Sequential Action Quantification & 2,103 \\
& State Transition Prediction & 2,875 \\
& Outcome Verification & 5,479 \\
& Recovery Reasoning & 2,213 \\
\bottomrule
\end{tabular}
}
\caption{Statistics of the current \projname benchmark release. MCQ / visual QA instances are counted from the main benchmark directory, while OpenQA items are deduplicated from the OpenQA evaluation pack, which contains a mirrored \texttt{data/openqa\_by\_path} copy. Multi-select items are verification-state questions requiring exact set matching.}
\label{tab:dataset_statistics}
\end{table}

\begin{figure}[htbp]
\centering
\includegraphics[width=1\linewidth]{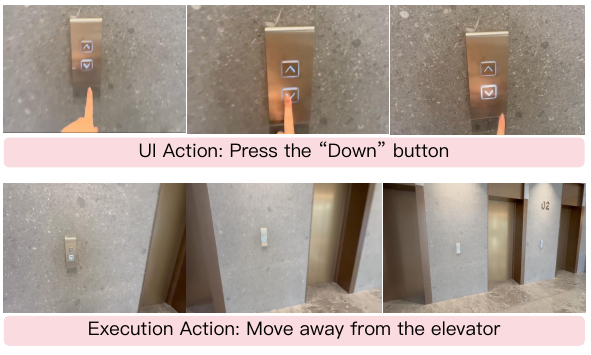}
\caption{Examples of two action categories. Left: UI Action. Right: Procedural Action.}

\label{fig:exampleForAction}
\end{figure}
\vspace{-3mm}

\subsection{Case Study Examples}
\label{app:case_studies}

We include representative MCQ and OpenQA examples to illustrate the main failure modes discussed in Section~\ref{sec:observations}. The MCQ cases highlight errors in state perception, temporal action reasoning, outcome verification, and recovery. The OpenQA cases show how free-form responses often become partial, contradictory, or too vague under rubric-based judging.

\begin{figure*}[htbp]
\centering
\includegraphics[width=1\linewidth]{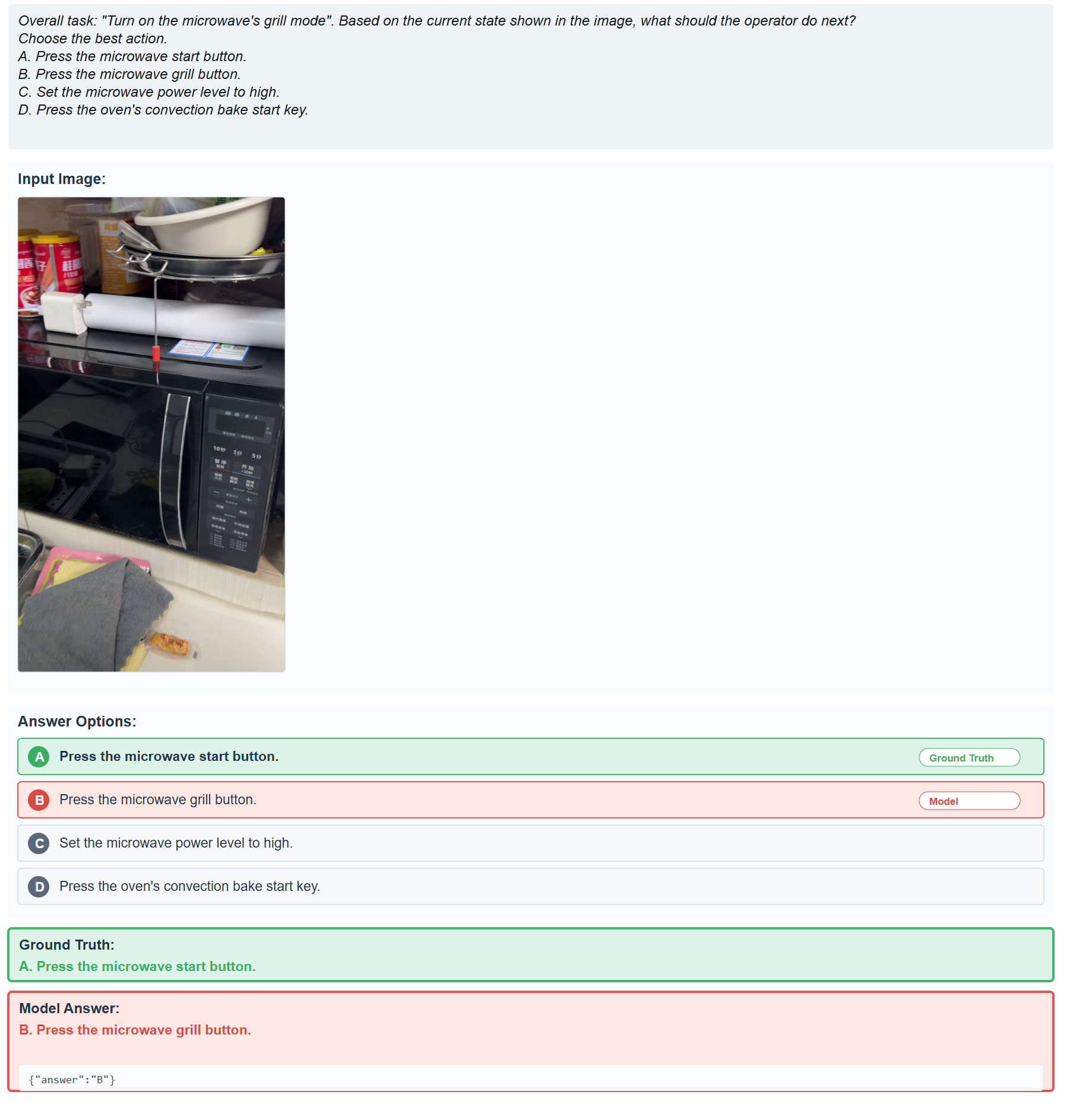}
\caption{Example (A): A Gemini 3.1 Pro failure on Task 3 / Action Generation in question format: image in question, text as option. The image shows a microwave interface after the grill mode has already been selected. The correct next action is therefore to press the start button to execute the heating process, but the model predicts the grill button instead. This error repeats a completed mode-selection step and shows that the model grounds the appliance category but misses the temporal state of the procedure.}
\label{fig:errorCase_gemini31pro_action}
\end{figure*}


\begin{figure*}[htbp]
\centering
\includegraphics[width=1\textwidth]{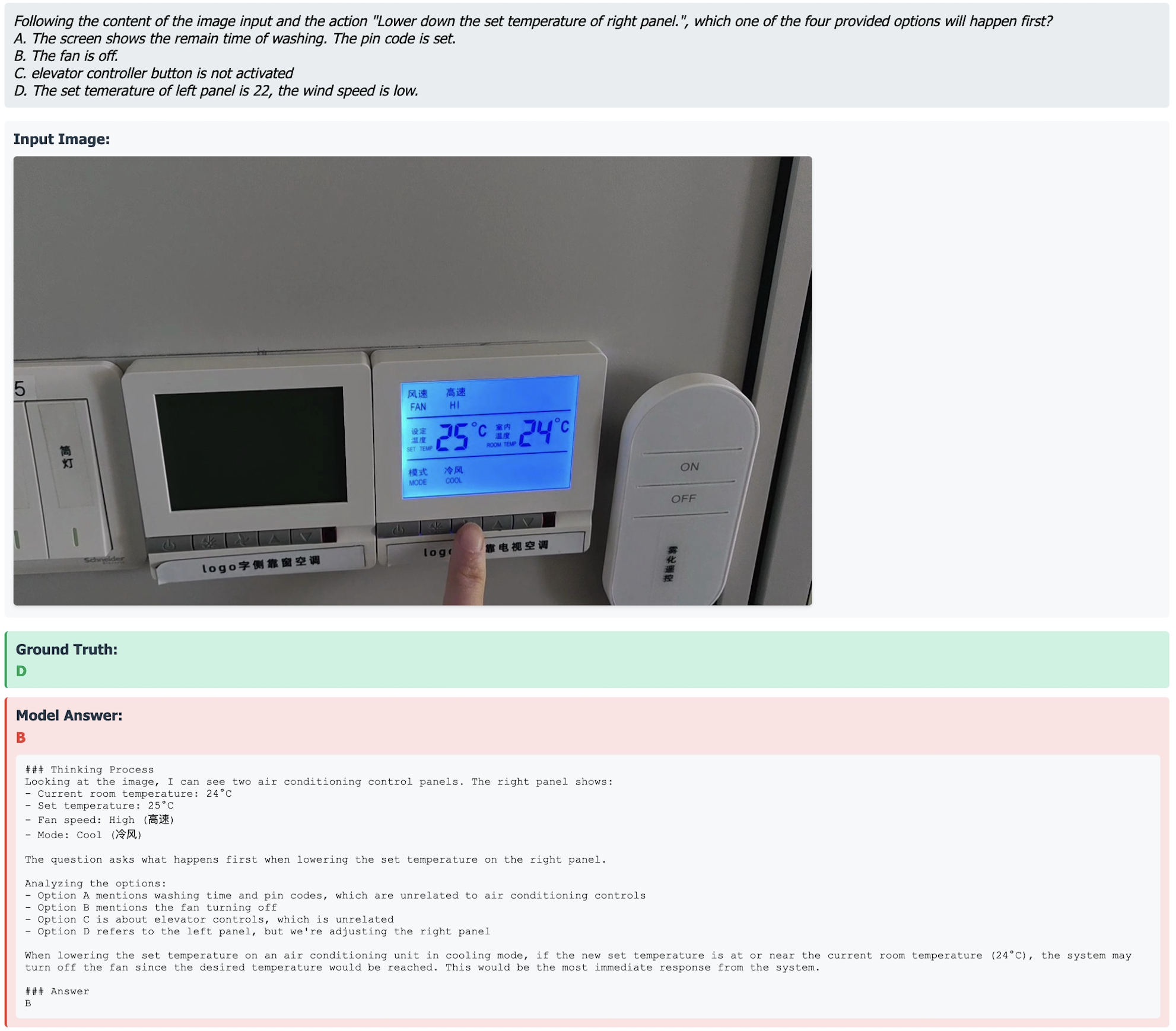}
\caption{Example (B): An example of Qwen3-VL-235B-Instruct on Task 4/ State Transition Prediction in question format: Image in question, text as option. The model fails to predict that the initial outcome of lowering the temperature on an air conditioner control is the change in the visible set temperature.}
\label{fig:errorCase_f}
\end{figure*}

\begin{figure*}[htbp]
\centering
\includegraphics[width=1\textwidth]{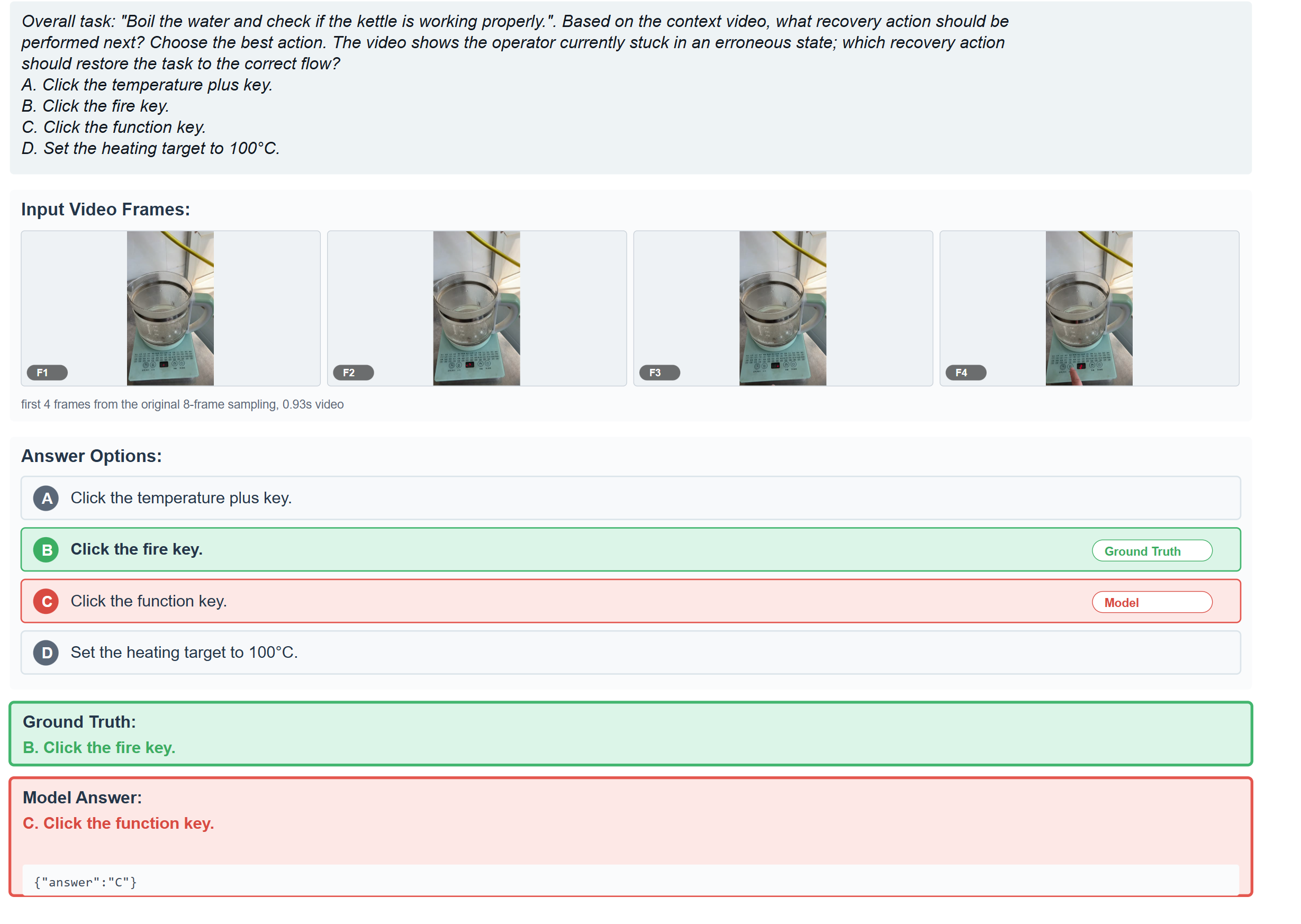}
\caption{
Example (C): A Gemini 3.1 Pro failure on Task 5 / Recovery Reasoning in the video-to-text multiple-choice setting.
The context video shows a kettle-control interaction in which the operator is in an erroneous or stalled state while trying to boil water and verify that the kettle is working properly.
The question asks which recovery action should restore the task to the intended execution flow.
The correct answer is \textbf{B}, clicking the fire key, because this action re-engages the heating process and recovers from the current incorrect state.
Gemini 3.1 Pro instead predicts \textbf{C}, clicking the function key.
Although this option is visually and semantically close to the correct control, it does not recover the task state required by the current procedure.
This case illustrates that strong multimodal models may recognize the appliance and nearby interface controls, yet still fail to identify the recovery-relevant action when procedural state, action history, and control semantics must be jointly reasoned over.
}
\label{fig:errorCase_gemini31pro_recovery}
\end{figure*}

\begin{figure*}[htbp]
\centering
\includegraphics[width=1\textwidth]{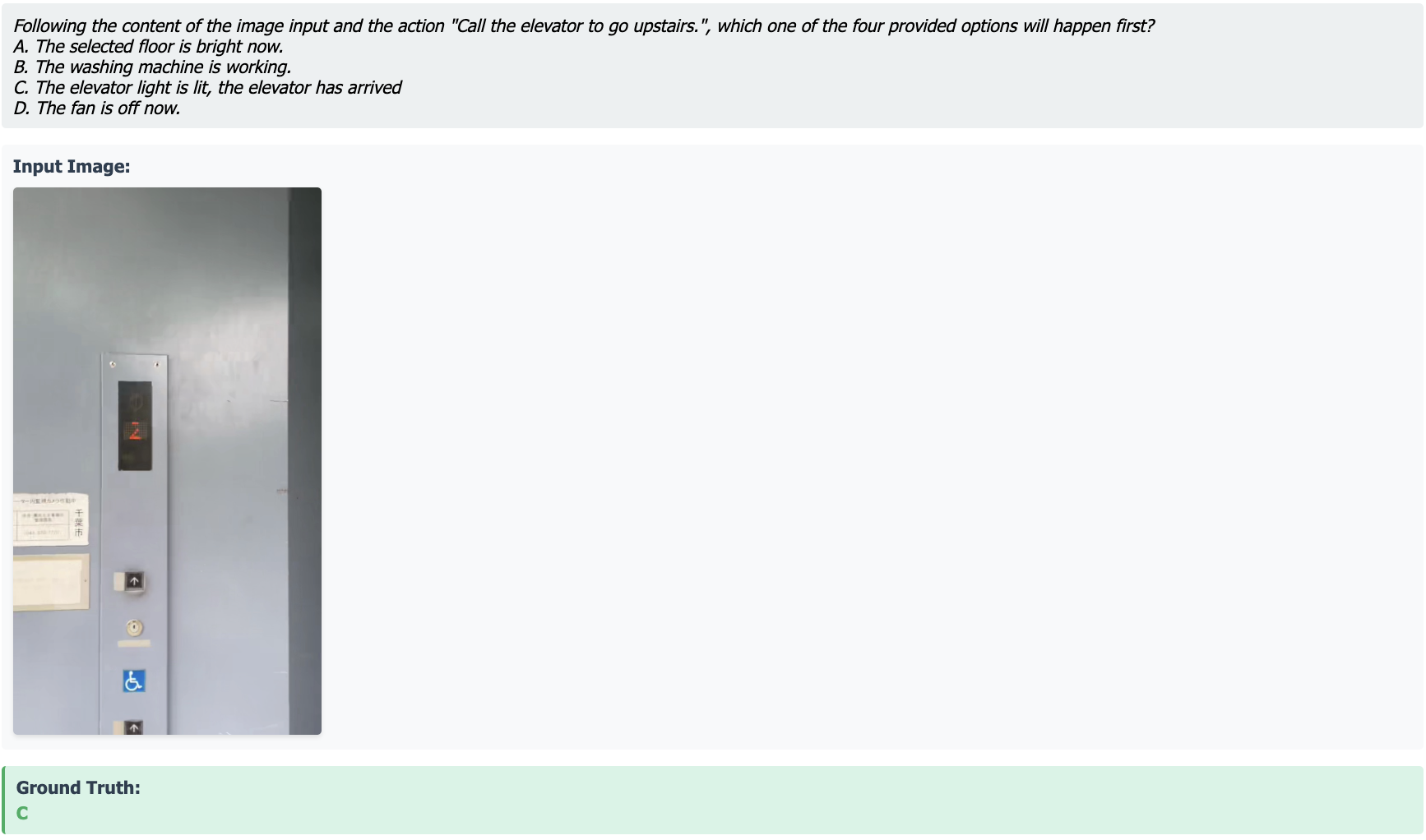}
\caption{Example (D) in question format: Image in question, text as option. Text choices may provide reasoning shortcuts for models.}
\label{fig:errorCase_g_textinput}
\end{figure*}

\begin{figure*}[htbp]
\centering
\includegraphics[width=0.95\textwidth]{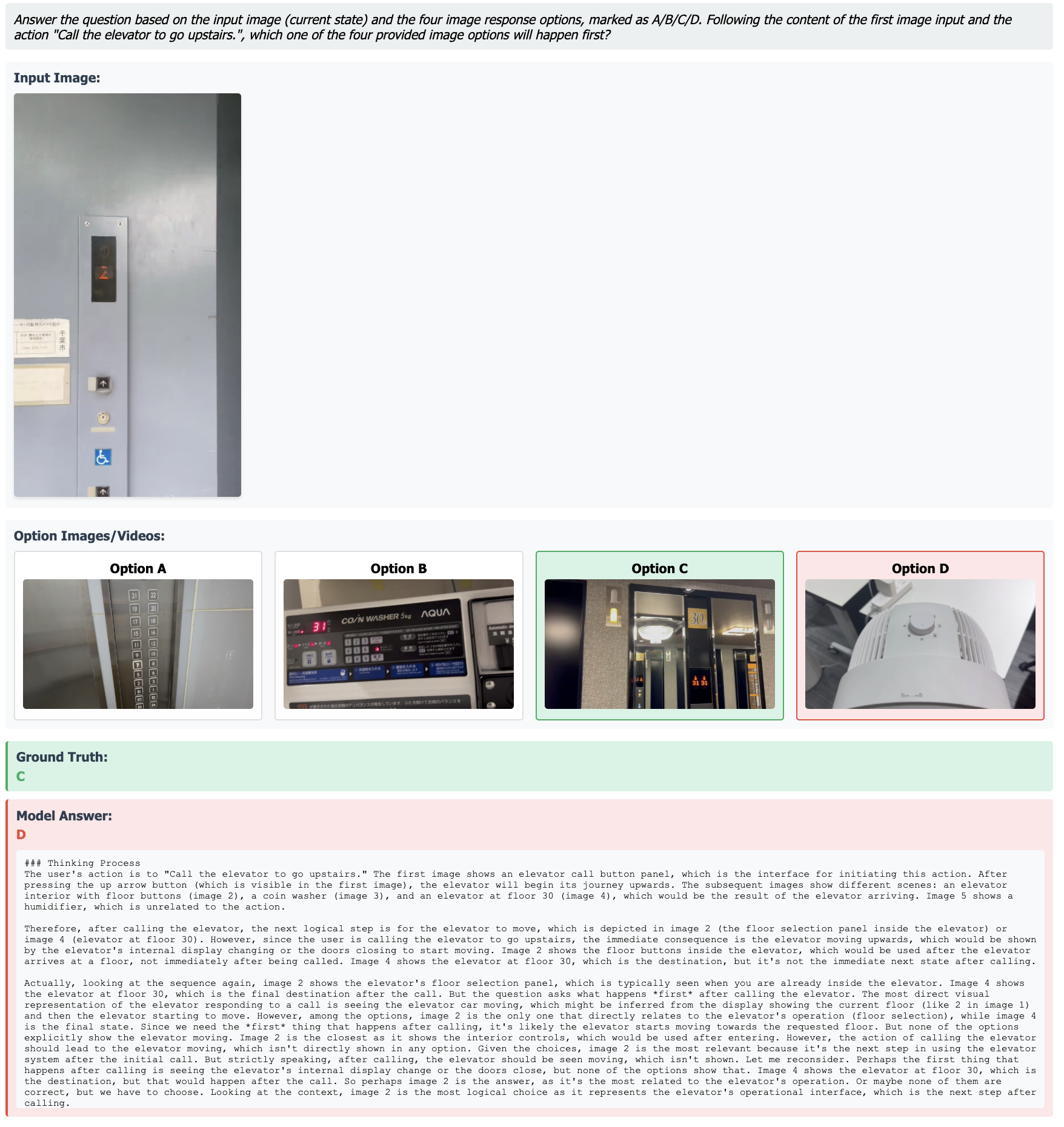}
\caption{Example (E): An example of Qwen3-VL-235B-Instruct on Task 4/ State Transition Prediction in question format: Image in question, images as option. The model fails to derive that "the elevator is arriving" from Option C, resulting in an incorrect answer.}
\label{fig:errorCase_g}
\end{figure*}

\begin{figure*}[htbp]
\centering
\includegraphics[width=0.95\textwidth]{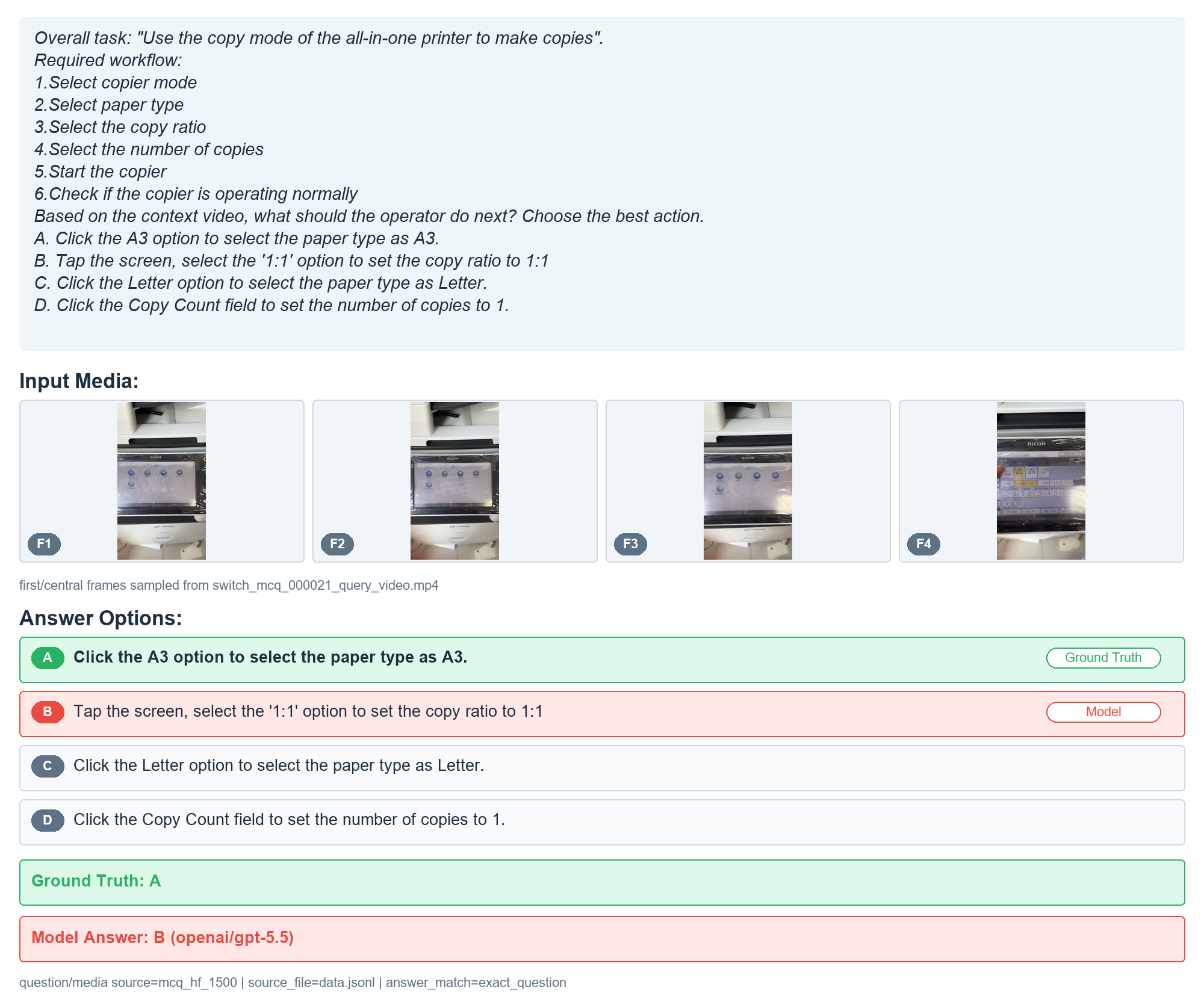}
\caption{Example (F):An example of GPT 5.5 on Task / Next Action Prediction in question format: Video in question, text as
  option. The model fails to follow the required copier workflow: after selecting the copier mode, the next action
  should be selecting the paper type as A3, but the model incorrectly predicts setting the copy ratio to 1:1. }
\label{fig:errorCase_g}
\end{figure*}

\begin{figure*}[htbp]
\centering
\includegraphics[width=0.95\textwidth]{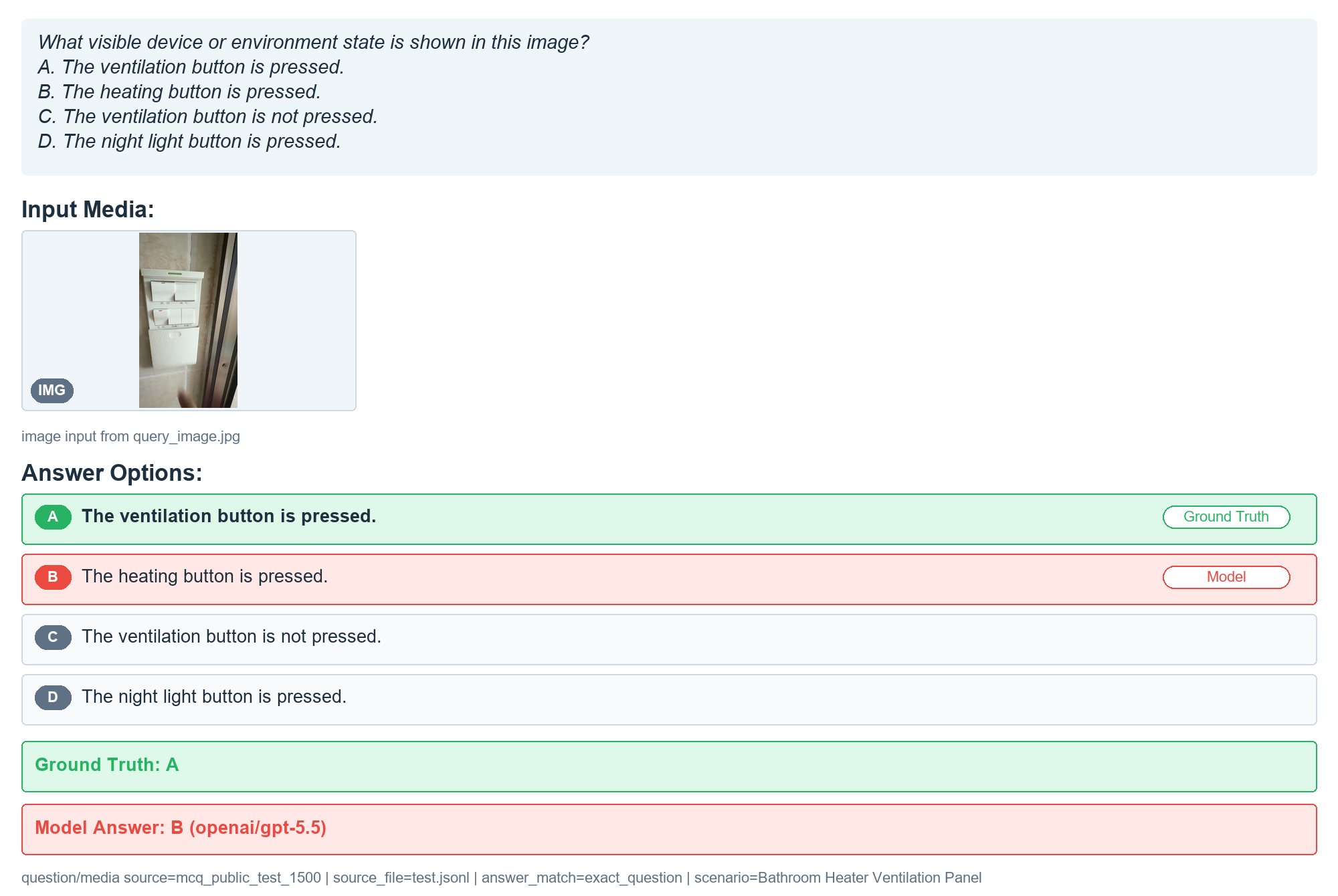}
\caption{Example (G):An example of GPT5.5 on Task / State Recognition in question format: Image in question, text as
  option. The model fails to identify the visible pressed button on the bathroom heater ventilation panel: the correct
  state is that the ventilation button is pressed, but the model incorrectly predicts that the heating button is
  pressed. }
\label{fig:errorCase_g}
\end{figure*}

\end{document}